\documentclass{article}
\usepackage{iclr2022_workshop,times}

\usepackage{amsmath,amsfonts,bm}

\def\eqref#1{equation~\ref{#1}}

\def\1{\bm{1}}

\DeclareMathAlphabet{\mathsfit}{\encodingdefault}{\sfdefault}{m}{sl}
\SetMathAlphabet{\mathsfit}{bold}{\encodingdefault}{\sfdefault}{bx}{n}

\usepackage{hyperref}
\usepackage{url}
\usepackage{graphicx}
\usepackage{wrapfig}
\usepackage{combelow}

\renewcommand{\vec}{\mathbf}

\title{Message passing all the way up}

\author{Petar Veli\v{c}kovi\'{c}\\
DeepMind / University of Cambridge\\
\texttt{petarv@deepmind.com}}

\iclrfinalcopy 
\begin{document}

\maketitle

\begin{abstract}
The message passing framework is the foundation of the immense success enjoyed by graph neural networks (GNNs) in recent years. In spite of its elegance, there exist many problems it provably cannot solve over given input graphs. This has led to a surge of research on going \emph{``beyond message passing''}, building GNNs which do not suffer from those limitations---a term which has become ubiquitous in regular discourse. However, have those methods truly moved beyond message passing? In this position paper, I argue about the dangers of using this term---especially when teaching graph representation learning to newcomers. I show that any function of interest we want to compute over graphs can, in all likelihood, be expressed using pairwise message passing -- just over a potentially \emph{modified} graph, and argue how most practical implementations subtly do this kind of trick anyway. Hoping to initiate a productive discussion, I propose replacing \emph{``beyond message passing''} with a more tame term, \textbf{``augmented message passing''}.
\end{abstract}

\section{Introduction}

In the span of only five years, graph neural networks (GNNs) have ascended from a niche of representation learning to one of its most coveted methods---enabling industrial and scientific applications that were not possible before. The growing list of applications includes recommender systems \citep{ying2018graph,hao2020p}, traffic prediction \citep{derrow2021eta}, chip design \citep{mirhoseini2021graph}, virtual drug screening \citep{stokes2020deep} and advances in pure mathematics \citep{davies2021advancing}, especially representation theory \citep{blundell2021towards}.

Most of these successes were propped up by the message passing framework \citep{gilmer2017neural}, where pairs of nodes exchange vector-based messages with one another in order to update their representations. However, fundamental limitations of this framework have been identified \citep{xu2018powerful,morris2019weisfeiler}---and it is unable to detect even the \emph{simplest} of substructures in graphs. For example, message passing neural networks provably cannot distinguish a 6-cycle  \includegraphics[width=3em,trim=0 0 165 0,clip]{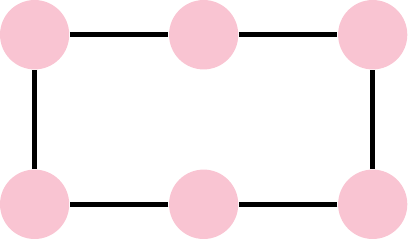} from two triangles  \includegraphics[width=3em,trim=165 0 0 0,clip]{tikz_WL.pdf} \citep{murphy2019relational,sato2021random}, and they are vulnerable to effects like oversmoothing \citep{li2018deeper} and oversquashing \citep{alon2020bottleneck}. These limitations have led to a surge in methods that aim to make structural recognition easier for GNNs, which has been undoubtedly one of the most active areas of graph representation learning in the recent few years. See \citet{maron2018invariant,murphy2019relational,chen2019equivalence,vignac2020building,morris2019weisfeiler,chen2020can,li2020distance,de2020natural} for just a handful of examples.

Collectively, these methods are known under many names, including \emph{higher-order GNNs} \citep{morris2019weisfeiler}, and \emph{beyond message passing}. In particular, the latter has become popularised by influential researchers in year-end summaries \citep{bronstein_2021,bronstein_2022} and was the central topic of debate at flagship workshops in the area \citep[GTRL'21]{cheng2021gtrl}.

By writing this position paper, I intend to initiate a serious debate on this term, which I consider to be harmful in many cases, \emph{especially} for newcomers entering the area. Namely, the very term ``beyond message passing'' apparently implies that the message passing primitive needs to be \emph{replaced} in order to break away from the inherent limitations mentioned above. But, how many of the methods that improve the expressive power of GNNs \emph{truly} moved beyond the message passing primitive?

I will make several arguments towards a \textbf{negative} answer. Firstly, in all likelihood, \emph{any} function of interest we want to compute over a graph \emph{can} be expressed using pairwise message passing---just over a potentially \emph{modified} graph structure. Beyond, many powerful GNNs are efficiently implementable \emph{exactly} using message passing. Even when a powerful GNN is easier or semantically clearer to implement with other primitives, interpreting it from the lens of message passing is possible, and can draw meaningful connections to research directions that are currently underappreciated. 

To be clear, I am absolutely in favour of research of expressive GNNs. However, teaching all of it as going \emph{``beyond message passing''} is, at best, imprecise and, at worst, harmful---as these methods often can, and often do, make efficient use of the message passing primitive.

Regarding the related \emph{higher-order GNN} term, I find its use to implicitly bind us to the Weisfeiler-Lehman hierarchy \citep{morris2021weisfeiler}, which is not always a meaningful measure of expressive power---see \citet{loukas2019graph,loukas2020hard, corso2020principal,barcelo2020logical} for examples. 

Taken all together, I propose that we coin a new term to describe recent progress in expressive GNNs: one that acknowledges the important role the message passing primitive will continue to play, while also realising we are surpassing its na\"{i}ve application. I propose \textbf{``augmented message passing''}, but am very much open to suggestions from the community.

\section{Message passing}

I will adopt the definition of message passing given by \citet{bronstein2021geometric}. Let a graph be a tuple of \emph{nodes} and \emph{edges}, $\mathcal{G}=(\mathcal{V}, \mathcal{E})$, with one-hop neighbourhoods defined as $\mathcal{N}_u = \{v\in\mathcal{V}\ |\ (v, u)\in\mathcal{E}\}$. Further, a node feature matrix ${\bf X}\in\mathbb{R}^{|\mathcal{V}|\times k}$ gives the features of node $u$ as $\vec{x}_u$; I omit edge- and graph-level features for clarity. A \emph{message passing} GNN over this graph is then executed as:
\begin{equation}
    \vec{h}_u = \phi\left(\vec{x}_u, \bigoplus\limits_{v\in\mathcal{N}_u} \psi(\vec{x}_u, \vec{x}_v)\right)
\end{equation}
where $\psi : \mathbb{R}^k\times\mathbb{R}^k\rightarrow\mathbb{R}^l$ is a \emph{message function}, $\phi : \mathbb{R}^k\times\mathbb{R}^l\rightarrow\mathbb{R}^m$ is a \emph{readout function}, and $\bigoplus$ is a permutation-invariant \emph{aggregation function} (such as $\sum$ or $\max$). Both $\psi$ and $\phi$ can be realised as MLPs, but many special cases exist, giving rise to, e.g., attentional GNNs \citep{velivckovic2017graph}.

\begin{wrapfigure}[10]{R}[0pt]{0.323\textwidth}
    \includegraphics[width=\linewidth]{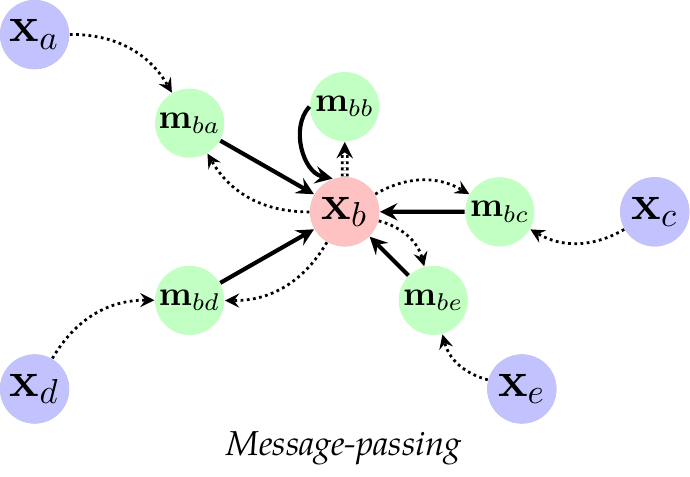}
\end{wrapfigure}

Message passing is a primitive that is propped up on \emph{pairwise communication}. This locality makes it scalable, but can also be its weakness; it is known that Equation 1 is incapable of recognising even simple substructures in graphs \citep{xu2018powerful}, and provably cannot model some processes over collections of more than two nodes \citep{neuhauser2021opinion}. Naturally, this spurred a plethora of research on improving upon Equation 1. I will unpack what that research actually did, and realise how it never truly escaped Equation 1, in terms of pairwise communication. The key is in allowing changes to the graph \emph{structure} (e.g. $\mathcal{V}$ or $\mathcal{N}_u$).

\paragraph{Simple example: \emph{Master nodes}}

To illustrate what is meant by changes to the graph structure, I will first discuss a simple case: that of handling \emph{graph-level features} (e.g. as in  \citet{battaglia2018relational}). This is useful either when graph-level context is given and should condition the computation, when shortening the graph's diameter through a bottleneck is desirable, or simply when tasks require graph-level decision-making.

The standard way in which such a feature is implemented \citep{gilmer2017neural} is by incorporating a \emph{master node}, $\mu$, connecting it with all other nodes, and then performing message passing as usual. Mathematically, $\mathcal{V}' = \mathcal{V}\cup\{\mu\}$, $\mathcal{N}'_u = \mathcal{N}_u\cup\{\mu\}$, and $\mathcal{N}'_\mu = \mathcal{V}$. 

It is now possible to do global-conditioned computation, and any two nodes are no more than two hops apart. Did we need to go \emph{beyond} message passing to do that? No---we \textbf{augmented} message passing to perform over a graph that has one extra node, and $O(\mathcal{V})$ new edges. All the while, the message passing primitive itself is unchanged\footnote{It is worth noting that often, we need to tag the added nodes or edges ($\mu$ in this case) with a special one-hot \emph{flag} feature, which would give the message function $\psi$ sufficient context to treat them separately.}. Such conclusions will repeat as we go on.

\section{Augmented message passing}

Broadly, augmented message passing techniques can be categorised into \emph{six} distinct types. 

\paragraph{Feature augmentation}

Since GNNs are unable to solve certain simple tasks on graphs, a very broad class of methods seek to (partially) precompute answers to those tasks, and feed the results as additional inputs. For example, it is well-known that even incorporating simple \emph{identifying} information such as a one-hot encoding \citep{murphy2019relational} or a random feature \citep{sato2021random} can assist GNNs in detecting patterns such as cycles. These can be further generalised to methods that literally \emph{count} subgraphs of interest \citep{bouritsas2020improving}, or provide the graph's Laplacian eigenvectors \citep{dwivedi2020generalization} as input. In all of the above, once the features are computed, the GNN computations proceed as before---hence, trivially expressible using message passing.

\paragraph{Message passing modulation}

While adding additional features can and does boost expressive power, it is arguably \emph{non-binding}: the model is not forced to use these features. Accordingly, the message function can be modulated to explicitly take into account these computations. One canonical example are directional graph networks \citep{beaini2021directional}, where Laplacian eigenvectors are used to define ``flows'' on a graph, which explicitly guide how various incoming messages are scaled. More recent proposals like LSPE \citep{dwivedi2021graph} explicitly include a message passing mechanism over the computed positional features. In all cases, while the message function, $\psi$, is modified, the blueprint of Equation 1 remains, and this case is also trivially an instance of message passing.

\paragraph{Graph rewiring}

The previous categories did not require moving away from message passing over exactly the given graph---at most, constraints needed to be placed on $\psi$. However, as the input graph is often noisy, missing, or suboptimal for the task, many methods \emph{modify} the edges of the input graph to compensate. Such \emph{graph rewiring} methods leave $\mathcal{V}$ unchanged, but make direct changes to $\mathcal{N}_u$.

One common method in this space is to operate over a \emph{fully connected graph}, i.e. setting $\mathcal{N}_u=\mathcal{V}$. This allows the model to rediscover the edges it needs, and has become quite popular in the context of \emph{graph Transformers} \citep{ying2021transformers,kreuzer2021rethinking,mialon2021graphit}, which allows for building GNNs that are able to ``win the hardware lottery'' \citep{hooker2021hardware}. Conveniently, the fully connected view also encompasses spectrally defined graph convolutions, such as the graph Fourier transform \citep{bruna2013spectral}. Nontrivial changes to $\mathcal{N}_u$, such as  multi-hop layers \citep{defferrard2016convolutional}, rewiring based on diffusion \citep{klicpera2019diffusion} or curvature \citep{topping2021understanding}, and subsampling \citep{hamilton2017inductive} are also supported. Lastly, the methods which dynamically alter the adjacency in a learnable fashion \citep{kipf2018neural,wang2019dynamic,kazi2020differentiable,velivckovic2020pointer} can also be classified under this umbrella. The underlying message function setup of Equation 1 is unchanged, hence these methods are still expressible using message passing.

\paragraph{Subgraph aggregation}

An extension of graph rewiring methods which has seen considerable interest lately concerns itself with learning graph representations by aggregating multiple \emph{subgraphs} at once---while being mindful of repeated nodes across those subgraphs; see \citet{papp2021dropgnn,cotta2021reconstruction,zhao2021stars,bevilacqua2021equivariant}. Such computations can be realised in the message passing framework by \emph{replicating} the nodes for every subgraph of interest, and \emph{connecting the copies} of every node together. Mathematically, if we are learning over $K$ subgraphs, let $\mathcal{V}'=\mathcal{V}\times\{1,2,\dots,K\}$ and, assuming that subgraph $i$'s edges define neighbourhoods $\mathcal{N}_u^{(i)}$, we can redefine neigbourhoods as follows: $\mathcal{N}'_{u, i} = \{(v, i)\ |\ v\in\mathcal{N}_u^{(i)}\}\cup\{(u, j)\ |\ j\in\{1,2,\dots,k\}\}$.

\paragraph{Substructure based methods}

As briefly mentioned, naturally-occurring phenomena are sometimes best described by interactions of \emph{groups} of entities---e.g., the \emph{functional groups} in a molecule can often strongly influence its properties \citep{duvenaud2015convolutional}. Accordingly, methods have been developed to support computing representations over junction trees \citep{fey2020hierarchical}, spectral signals \citep{stachenfeld2020graph}, simplicial complexes \citep{bodnar2021simplicial}, cellular complexes \citep{bodnar2021weisfeiler}, or general $k$-tuples of nodes \citep{morris2019weisfeiler,morris2020weisfeiler} and hypergraphs \citep{huang2021unignn,chien2021you,georgiev2022heat}. Whenever the groups of interest are greater than two nodes, the expressivity of pairwise message passing can be brought into question. Indeed, over the \emph{original} graph, it can be provably impossible to use pairwise messaging to simulate such interactions \citep{neuhauser2021opinion}. But what if we \emph{modify} the graph structure?

Indeed, we can; and very often, such methods will explicitly call out the use of message passing to achieve their objective \citep{fey2020hierarchical,stachenfeld2020graph}. The trick is to create new nodes for every substructure that we want to model, and appropriately connecting them to their constituent nodes. How easy this is to do depends on whether the functions of interest are \emph{permutation invariant}.

If yes, all we have to do is establish bidirectional edges between the constituent nodes and the corresponding ``substructure node''. Mathematically, we assume that we have $K$ substructures, $\mathcal{S}_1, \mathcal{S}_2,\dots,\mathcal{S}_K$ each operating in a permutation invariant way over its constituent nodes ($\mathcal{S}_i\subseteq\mathcal{V}$). Then, we augment the graph by creating new substructure nodes: $\mathcal{V}'=\mathcal{V}\cup\{\mu_1,\mu_2,\dots,\mu_K\}$, and modifying the neighbourhoods to connect every constituent to its substructure(s): $\mathcal{N}'_u=\mathcal{N}_u\cup\{\mu_i\ |\ u\in\mathcal{S}_i\}$, and $\mathcal{N}'_{\mu_i} = \mathcal{S}_i$. Note that this is a more general case of the master node approach of Section 2, which can be seen as having just one substructure, $\mathcal{S}_1=\mathcal{V}$.

To prove that \emph{any} permutation invariant function over substructures can be represented in this way, it is sufficient to realise that, when computing substructure representations $\vec{h}_{\mu_i}$, Equation 1 implements a \emph{Deep Sets} model over the nodes in $\mathcal{S}_i$ \citep{zaheer2017deep}. Under certain conditions on the message function hyperparameters, it is known that \emph{any} permutation invariant function over $\mathcal{S}_i$ \emph{must} be expressible in this form \citep{wagstaff2019limitations}---hence, this construction is universal.

When interactions within a substructure are permutation sensitive, a more intricate gadget is required which, due to space constraints, I fully derive in Appendix A. In a nutshell, we can either create $O(\mathcal{S}_i)$ new nodes to process nodes' features \emph{one at a time} acccording to the permutation (not unlike a long short-term memory \citep{hochreiter1997long}) or using carefully constructed message functions to materialise a \emph{concatenation} of the inputs which respects the permutation. 

In both cases, we have demonstrated an equivalent architecture using only the pairwise message passing primitive, which is often the way such methods are implemented anyway.

\paragraph{General equivariant GNNs}

While previous sections all sought to identify a \emph{specific} computation or motif that can improve GNN expressivity, a converse approach is to characterise \emph{all} possible linear permutation-equivariant layers over an input graph, and use them as a basis for building equivariant GNNs \citep{maron2018invariant}. A similar analysis over image data reveals that there is exactly one type of linear translation equivariant layer over images: the convolution \citep{bronstein2021geometric}.

Using this framework, \citet{maron2018invariant} discover a basis of two linear invariant and 15 linear equivariant layers for the setting with edge-valued inputs (for data defined over $k$-tuples of nodes, the dimension is defined by the $k$-th and $2k$-th Bell numbers). These 15 layers effectively allow for recombining information across \emph{pairs of edges} while respecting the graph structure's symmetries. Therefore, they are represented as matrices in $\mathbb{R}^{\mathcal{V}^2\times\mathcal{V}^2}$, which get multiplied with edge features.

Despite the clearly tensorial semantics of such an approach, it can still be represented in the language of pairwise message passing: any square matrix multiplication operation represents a convolutional GNN instance over the graph implied by that matrix's nonzero entries \citep{bronstein2021geometric}. The key difference is that in this case the messages are passed over \emph{edges} rather than nodes, and therefore we need to invent new nodes corresponding to edges, and connect them accordingly!

Mathematically, assume we want to multiply with a basis matrix ${\bf B}\in\mathbb{R}^{\mathcal{V}^2\times\mathcal{V}^2}$. Inventing new edge-based nodes corresponds to $\mathcal{V}'=\mathcal{V}\cup\{e_{uv}\ |\ u, v\in\mathcal{V}\}$. Then, we need to connect these edges to the nodes incident to them, and also to other edges, whenever the entries of $\bf B$ mandate it. Hence the neighbourhoods update as follows: $\mathcal{N}'_u=\mathcal{N}_u\cup\{e_{ab}\ |\ a=u\vee b=u\}$ for nodes, and $\mathcal{N}'_{e_{uv}} = \{u, v\}\cup\{e_{ab}\ |\ {\bf B}_{(a, b),(u, v)} \neq 0\}$ for edges. Similar gadgets (potentially ``tensored up'' to $k$-tuple nodes) would also apply for follow-up work, including but not limited to \citet{maron2019provably,keriven2019universal,albooyeh2019incidence,azizian2020characterizing}.

In this particular case, the semantics of message passing do not ``line up'' very nicely with the tensorial approach. Still, expressing the edges as nodes reveals connections to several proposals that \emph{do} directly augment message passing, especially over the related concept of a \emph{line graph} \citep{monti2018dual,cai2021line}. Hence, even when the conversion to message passing is not the most practical, it can still be a useful exercise that reveals surprising connections between seemingly unrelated proposals, and stimulates future research. 

\subsubsection*{Acknowledgments}
In many ways, this was a step in the unknown for me. Writing a position paper by myself, and exposing it to the critique of my peers, is something quite daunting, no matter how much I believe in the idea! I wish to give a deep thank-you to everyone who reviewed, proofread and critiqued drafts of this paper, making the whole journey a lot less scary: Pete Battaglia, Michael Bronstein, Andreea Deac, Jonny Godwin, Thomas Kipf, Andreas Loukas, Haggai Maron and Christopher Morris.

This paper was spearheaded through the Master's course on graph representation learning at Cambridge, which I am currently co-teaching with Pietro Li\`{o}. The need to write it arose directly from interactions with our great students, realising the direct negative impacts the term ``beyond message passing'' can have. These concepts were solidified after fantastic and intense discussions with the teaching team: Cris Bodnar, Iulia Du\cb{t}\u{a}, Dobrik Georgiev, Charlie Harris, Chaitanya Joshi, Paul Scherer and Ramon Vi\~{n}as. Thanks are very much due to all of them!

\bibliography{iclr2022_workshop}
\bibliographystyle{iclr2022_workshop}

\newpage
\appendix

\section{Graph modulation for permutation-sensitive substructures}

Let's assume that we want to model a function $f(\vec{x}_1, \vec{x}_2, \dots, \vec{x}_k)$ over $k$-tuples of nodes, and that it is permutation sensitive in the first $n$ arguments, and permutation invariant in the remaining $m$ (s.t. $n+m=k$). Just like before, we can fit the $m$ invariant parameters in the same way as in the main paper (inventing a new ``master node'', $\mu$, that exchanges messages with those $m$ nodes). This gives us a modified function, re-using Equation 1: \begin{equation}f(\vec{x}_1, \vec{x}_2, \dots, \vec{x}_k) = g\left(\vec{x}_1, \dots, \vec{x}_n, \phi_1\left(\vec{x}_\mu, \bigoplus\limits_{l=1}^m \psi_1(\vec{x}_\mu, \vec{x}_{n+l})\right)\right)\end{equation}
where we can set the master node's features, $\vec{x}_\mu$ as a zero-vector initially, if none are available.

Now, we need to process all of these $n+1$ inputs to $g$ in a way that is order-dependent. The simplest way to do so is to create a new node which would use ``concat-aggregation'', but it is unclear whether this is explicitly supported by Equation 1. There do exist ways to materialise this concatenation by preparing message functions that copy each of the $\vec{x}_i$ vectors into a separate ``slot'' of the result vector, then applying the sum aggregation. Then the concatenated node can send its own features as a message to a node storing the result.

If we want to avoid explicitly materialising concatenations, it is also possible to use Turing-complete recurrent models such as the LSTM \citep{hochreiter1997long} to gradually process the inputs to $g$ one at a time, storing intermediate results $\pmb{\lambda}_i$ as we go along ($i\in\{1, 2, \dots, n+1\}$).

Initialising $\pmb{\lambda}_i=\vec{0}$ (or making it learnable), we then update each of them as follows:
\begin{equation}
    \pmb{\lambda}'_i= \begin{cases}\phi_2\left(\pmb{\lambda}_i, \psi_2(\pmb{\lambda}_{i}, \pmb{\lambda}_{i-1}) + \psi_2(\pmb{\lambda}_{i}, \vec{x}_i)\right) & i \leq n\\
    \phi_2\left(\pmb{\lambda}_i, \psi_2(\pmb{\lambda}_{i}, \pmb{\lambda}_{i-1}) + \psi_2(\pmb{\lambda}_{i}, \vec{x}_\mu)\right) & i = n + 1\end{cases}
\end{equation}
which aligns with the message passing framework of Equation 1 (setting $\mathcal{N}_{\lambda_i} = \{\lambda_{i-1}, i\}$, and specially $\mathcal{N}_{\lambda_{n+1}} = \{\lambda_{n}, \mu\}$). Further, these equations align with recurrent neural networks (e.g. we can recover LSTMs for a special choice of $\phi_2$ and $\psi_2$). Once $n+1$ steps of such a model have been performed, sufficient time has passed for all the features to propagate, and we may use $\pmb{\lambda}_{n+1}$ as the final representation of our target function $f$, potentially feeding its outputs back to relevant nodes.

\end{document}